\documentclass[10pt,twocolumn,letterpaper]{article}

\usepackage[pagenumbers]{cvpr} 
\usepackage{booktabs}
\usepackage{multirow}
\usepackage{tabularx}
\usepackage{makecell}
\usepackage{booktabs}
\usepackage{multirow}
\usepackage{bbding}
\usepackage{color}
\usepackage{url}
\usepackage{booktabs}
\definecolor{green_c}{RGB}{0, 209, 0} 
\definecolor{red}{RGB}{255,0,0} 
\usepackage[marginal]{footmisc}
\usepackage{float}
\usepackage[dvipsnames]{xcolor}

\definecolor{cvprblue}{rgb}{0.21,0.49,0.74}
\usepackage[pagebackref,breaklinks,colorlinks,citecolor=cvprblue]{hyperref}

\def\paperID{717} 
\def\confName{CVPR}
\def\confYear{2024}

\title{SeeSR: Towards Semantics-Aware Real-World Image Super-Resolution}

\author{Rongyuan Wu$^{1,2}$,  Tao Yang$^{3}$,  Lingchen Sun$^{1,2}$,  Zhengqiang Zhang$^{1,2}$,  Shuai Li$^{1,2}$, Lei Zhang$^{1,2,}$\thanks{Corresponding author. This work is supported by the Hong Kong RGC RIF grant (R5001-18) and the PolyU-OPPO Joint Innovation Lab.} \\
{$^{1}$The Hong Kong Polytechnic University \qquad $^{2}$OPPO Research Institute \qquad $^{3}$ByteDance Inc.}
 \\
{\small \{rong-yuan.wu, ling-chen.sun, zhengqiang.zhang, novak.li\}@connect.polyu.hk} \  \\
{\small \ yangtao9009@gmail.com, \ cslzhang@comp.polyu.edu.hk}
\\
}

\begin{document}

\maketitle
\thispagestyle{empty}
\begin{abstract}
Owe to the powerful generative priors, the pre-trained text-to-image (T2I) diffusion models have become increasingly popular in solving the real-world image super-resolution problem. However, as a consequence of the heavy quality degradation of input low-resolution (LR) images, the destruction of local structures can lead to ambiguous image semantics. As a result, the content of reproduced high-resolution image may have semantic errors, deteriorating the super-resolution performance. To address this issue, we present a semantics-aware approach to better preserve the semantic fidelity of generative real-world image super-resolution. First, we train a degradation-aware prompt extractor, which can generate accurate soft and hard semantic prompts even under strong degradation. The hard semantic prompts refer to the image tags, aiming to enhance the local perception ability of the T2I model, while the soft semantic prompts compensate for the hard ones to provide additional representation information. These semantic prompts encourage the T2I model to generate detailed and semantically accurate results. Furthermore, during the inference process, we integrate the LR images into the initial sampling noise to mitigate the diffusion model's tendency to generate excessive random details. The experiments show that our method can reproduce more realistic image details and hold better the semantics. The source code of our method can be found at \href{https://github.com/cswry/SeeSR}{https://github.com/cswry/SeeSR}.
\end{abstract}

\section{Introduction}
\label{sec:intro}

Images inevitably undergo degradation due to factors such as subpar imaging devices, unfavorable capturing environments, transmission losses, \etc. This degradation manifests in various forms, including low-resolution, blurriness and noise. Image super-resolution (ISR) aims to reconstruct a high-resolution (HR) image from the given low-resolution (LR) input. Traditionally, researchers investigate the ISR problem by assuming simple and known image degradations (\eg, bicubic downsampling), and developed many successful models \cite{dong2014learning,rcan, chen2021pre, liang2021swinir, zhang2022efficient,chen2023activating, ma2023text, zhang2023tmp, sun2023perception}. However, these methods often yield over-smoothed outcomes due to their fidelity-focused learning objectives. To enhance visual perception, generative adversarial networks (GANs) \cite{goodfellow2020generative} have been adopted to solve the ISR problem \cite{wang2018esrgan}. By using the adversarial loss in training, the ISR models can be supervised to generate perceptually realistic details, yet there can be much visual artifacts.

Despite the remarkable advancements, when applying the above mentioned models to real-world LR images, whose degradations are much more complex and even unknown, the output HR images can have low visual quality with many artifacts. This is mainly caused by the domain gap between the synthetic training data and the real-world test data. The goal of real-world ISR (Real-ISR) is to reproduce a perceptually realistic HR image from its LR observation with complex and unknown degradation. To this end, some researchers proposed to collect real-world LR-HR image pairs using long-short camera focal lens \cite{realsr, drealsr}. Another more cost-effective way is to simulate the complex real-world image degradation process using random combinations of basic degradation operations. The representative work along this line include BSRGAN \cite{zhang2021designing}, Real-ESRGAN \cite{wang2021real} and their variants \cite{liang2022details, liang2022efficient, chen2023human, xie2023desra}. With the abundant amount of more realistic synthetic training pairs, the GAN-based Real-ISR methods can generate more authentic details. However, they still tend to introduce many unpleasant visual artifacts due to the unstable adversarial training. LDL \cite{liang2022details} can suppress much the visual artifacts by detecting the problematic pixels using local image statistics. Unfortunately, it is not able to generate additional details.  

Recently, denoising diffusion probabilistic models (DDPMs) \cite{ho2020denoising} have exhibited remarkable performance in the realm of image generation, gradually emerging as successors to GANs in various downstream tasks \cite{saharia2022photorealistic,rombach2022high}. Some researchers \cite{kawar2022denoising,wang2022zero} have leveraged pretrained DDPMs to effectively tackle the inverse image restoration problems. However, their application to the challenging Real-ISR scenarios is hindered by the assumptions of known linear degradation model. Considering that the large-scale pretrained text-to-image (T2I) models \cite{rombach2022high, saharia2022photorealistic}, which are trained on a dataset exceeding 5 billion image-text pairs, encompass more potent natural image priors, some methods have recently emerged to harness their potentials to address the Real-ISR problem, including StableSR \cite{wang2023exploiting}, PASD \cite{yang2023pixel} and DiffBIR \cite{lin2023diffbir}. 
These diffusion prior based Real-ISR methods have demonstrated highly promising capability to generate realistic image details; however, they still have some limitations. StableSR \cite{wang2023exploiting} and DiffBIR \cite{lin2023diffbir} solely rely on input LR images as control signals, overlooking the role of semantic text information in the pretrained T2I models. PASD \cite{yang2023pixel} attempts to utilize off-the-shelf high-level models to extract semantic prompts as additional control conditions for the T2I model. However, it encounters difficulties when dealing with scenes containing a variety of objects or severely degraded images. 

In this work, we investigate in-depth the problem that how to extract more effective semantic prompts to harness the generative potential of pretrained T2I models so that better Real-ISR results can be obtained. By analyzing the effects of different types of semantic prompts on the Real-ISR outcomes, we conclude two major criteria. Firstly, the prompt should  cover as many objects in the scene as possible, helping the T2I model to understand different local regions of the LR image. Secondly, the prompt should be degradation-aware to avoid erroneous semantic restoration results. (Please refer to Section \ref{sec:method} for more discussions.) While the prompt extractor undergoes low-level data augmentation during training \cite{hendrycks2019augmix}, there still exists much gap between this augmentation and real-world degradation. Hence, it is not suitable to directly extract semantic prompts from real-world LR inputs.

Based on the aforementioned criteria, we present a \textbf{Se}mantic-awar\textbf{e} \textbf{SR} (\textbf{SeeSR}) approach, which utilizes high-quality semantic prompts to enhance the generative capacity of pretrained T2I models for Real-ISR. SeeSR consists of two stages. In the first stage, the semantic prompt extractor is fine-tuned to acquire degradation-aware capabilities. This enables it to  extract accurate semantic information from LR images as soft and hard prompts. In the second stage, the pristine semantic prompts collaborate with LR images to exert precise control over the T2I model, facilitating the generation of rich and semantically correct details. Moreover, during inference stage, we incorporate the LR image into the initial sampling noise to alleviate the diffusion model's propensity for generating excessive random details. 
Our extensive experiments demonstrate the superior realistic detail generation performance of SeeSR while preserving well the image semantics of Real-ISR outputs.

\section{Related Work}
\label{sec:related}
\vspace{-3mm}
\begin{figure*}[h]
  \centering
  \includegraphics[scale=0.68]{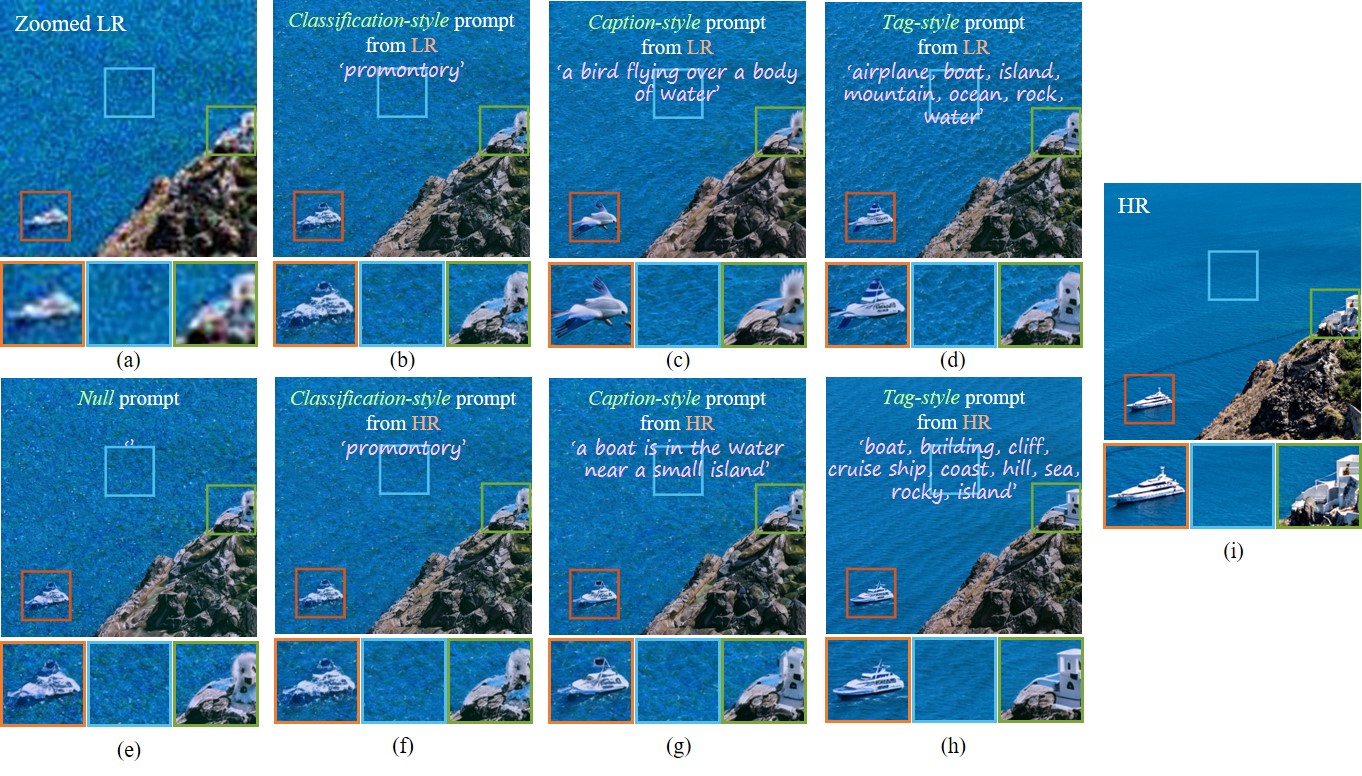}
  \caption{The comparison of different styles of prompts and their corresponding Real-ISR results with PASD \cite{yang2023pixel}. (a) Input LR image. (b)-(d) show the extracted classification-style, caption-style and tag-style prompts from LR image and the corresponding Real-ISR results. (e) Null prompt and its corresponding Real-ISR result. (f)-(h) show the extracted classification-style, caption-style and tag-style prompts from HR image and their corresponding Real-ISR results. (i) HR image.}
  \label{fig:observation}
  \vspace{-1.0em}
\end{figure*}

\noindent
\textbf{GAN-based Real-ISR.} Starting from SRCNN \cite{dong2014learning}, deep learning based ISR has become prevalent. A variety of methods focusing on model design have been proposed \cite{edsr,rcan,rdn,san,chen2021pre,liang2021swinir,zhang2022efficient,chen2023activating, dat} to improve the accuracy of ISR reconstruction. However, most of these methods assume simple and known degradations such as bicubic downsampling, limiting their effectiveness when dealing with complex and unknown degradations in real world.
Recent advancements in Real-ISR have explored more complex degradation models to approximate the real-world degradations. Specifically, BSRGAN \cite{zhang2021designing} introduces a randomly shuffled degradation modeling strategy, while Real-ESRGAN \cite{wang2021real} employs a high-order degradation modeling process. Using the training samples with more realistic degradations, both BSRGAN and Real-ESRGAN utilize GANs \cite{goodfellow2020generative} to reconstruct desired HR images. While generating more perceptually realistic details, the training of GANs is unstable and Real-ISR outputs often suffer from unnatural visual artifacts. Many following works such as LDL \cite{liang2022details} and DeSRA \cite{xie2023desra} can suppress much the artifacts, yet they are difficult to generate more natural details. 

\noindent
\textbf{Diffusion Probabilistic Models.}
Inspired by the non-equilibrium thermodynamics theory \cite{jarzynski1997equilibrium} and sequential MonteCarlo \cite{neal2001annealed}, Sohl-Dickstein \etal \cite{sohl2015deep} proposed the diffusion model to model complex datasets.  Subsequently, a series of fruitful endeavors \cite{ho2020denoising, song2020score, dhariwal2021diffusion} have been made to apply diffusion models in the realm of image generation, especially since the development of DDPM \cite{ho2020denoising}. Rombach \etal \cite{rombach2022high} expanded the training of DDPMs to the latent space, greatly facilitating the development of large-scale pretrained text-to-image (T2I) diffusion models such as stable diffusion (SD) \cite{sd} and Imagen \cite{imagen}. It has been demonstrated that T2I diffusion priors are powerful in image editing \cite{mou2023t2i, zhang2023adding}, video generation \cite{singer2022make, wu2023tune}, 3D content generation \cite{wang2023prolificdreamer, lin2023magic3d, wang2023prolificdreamer}, \etc.  

\noindent
\textbf{Diffusion Prior based Real-ISR.}
Early attempts \cite{sr3, kawar2022denoising, wang2022zero} using DDPMs to address the ISR problem are mostly assuming simple downsampling degradation. However, such an assumption of known linear image degradation restricts their practical application in complex scenarios like Real-ISR. Recently, some researchers \cite{wang2023exploiting, yang2023pixel, lin2023diffbir, sun2023improving} have employed powerful pretrained T2I models such as SD \cite{sd} to tackle the real-ISR problem. Having been trained on billions of image-text pairs, these models can perceive strong image priors for tackling Real-ISR challenges. StableSR \cite{wang2023exploiting} achieves this goal by training a time-aware encoder to fine-tune the SD model and employing feature warping to balance between fidelity and perceptual quality. DiffBIR \cite{lin2023diffbir} adopts a two-stage strategy to tackle the Real-ISR problem. It first reconstructs the image as an initial estimation, and then utilizes the SD prior to enhance image details. 

The aforementioned methods solely rely on images as conditions to activate the generative capability of the T2I model. In contrast, PASD \cite{yang2023pixel} goes further by utilizing off-the-shelf high-level models (\ie, ResNet \cite{he2016deep}, Yolo \cite{redmon2016you} and BLIP \cite{li2022blip}) to extract semantic information to guide the diffusion process, stimulating more generative capacity of the T2I model. However, ResNet and Yolo have limited object recognition ability, leading to a diminished recall rate. The captions generated by BLIP struggle to comprehensively describe the semantic information in images, particularly in scenes with a rich diversity of objects. Therefore, how to introduce prompts to more  effectively elicit the potential of pretrained T2I models in assisting Real-ISR needs deep investigation, which is the goal of this work.



\section{Methodology}
\subsection{Motivation and Framework Overview}
\label{sec:method}

\textbf{Motivation.} To unleash the generative potential of pretrained T2I model while avoiding semantic distortion in Real-ISR outputs, we investigate the use of three representative styles of semantic prompts, including \textit{classification-style}, \textit{caption-style} and \textit{tag-style}. In specific, we use the methods in \cite{he2016deep} , \cite{li2022blip} and \cite{2023ram} to extract classification-style, caption-style and tag-style prompts, respectively. 

The classification-style prompt provides only one category label for the entire image, which is robust to image degradation due to its global view. However, such kind of prompts lack the ability to provide semantic support of local objects, particularly in scenes containing multiple entities. As shown in Figs. \ref{fig:observation}(b) and \ref{fig:observation}(f), by using the classification-style prompts extracted from the LR and the HR images, the Real-ISR results are almost indistinguishable from that obtained by using the null prompt (see Fig. \ref{fig:observation}(e)). 

The caption-style prompt provides a sentence to describe the corresponding image, offering richer information compared to the classification-style prompt. However, it still has two shortcomings. Firstly, the redundant prepositions and adverbs in this type of prompt may scatter the attention of T2I models towards degraded objects \cite{prompt2prompt}. Secondly, it is prone to semantic errors due to the influence of degradation in LR images. As shown in Fig. \ref{fig:observation}(c), the T2I model mistakenly reconstructs a bird instead of a ship due to the incorrect caption extracted from the LR image. 

The tag-style prompt provides category information for all objects in the image, offering a more detailed description of the entities compared to caption-style prompt. Even without providing object location information, it is found that the T2I model can align the semantic prompts with the corresponding regions in the image due to its underlying semantic segmentation capability \cite{zhang2023tale}. Unfortunately, similar to the captioning models, the tagging models are also susceptible to image degradations, resulting in erroneous semantic cues and semantic distortion in the reconstructed results. As shown in Fig. \ref{fig:observation}(d), the wrong semantic prompt ``airplane" leads to distorted reconstruction of the ship. 

We summarize the characteristics of different styles of prompts in Table \ref{tab:prompt_style}. This motivates us that if we can adapt the tag-style prompt to be degradation-ware, then it may help the T2I models generate high-quality Real-ISR outputs while preserving correct image semantics. 

\begin{figure*}[htbp]
  \centering
  \includegraphics[scale=0.50]{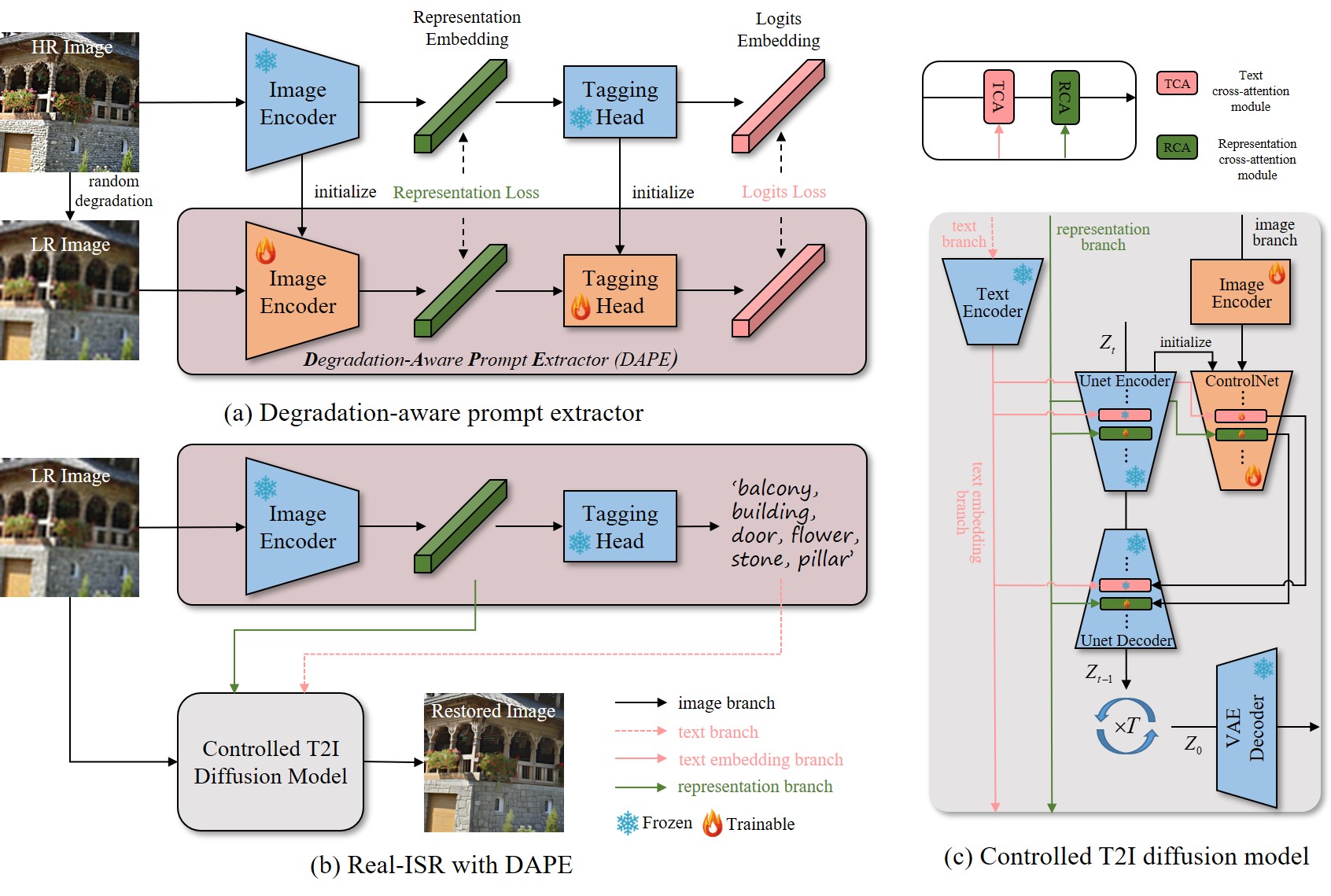}
  \caption{Overview of SeeSR. (a) In the first stage, we train a degradation-aware prompt extractor (DAPE), which is initialized from a tag model. DAPE is trained to align the encoding of the degraded LR image to the encoding of the corresponding HR image by a tag model (\eg, RAM \cite{2023ram} in our work), enabling DAPE the degradation-awareness. (b) In the second stage, the well-trained DAPE provides both soft prompts (representation embedding) and hard prompts (tagging text), which are combined with the LR image to control a pretrained T2I model (\eg, SD \cite{rombach2022high} in our work). The detailed structure of the controlled T2I diffusion model is shown in (c).}
  \label{fig:method}
  \vspace{-1.0em}
\end{figure*}

\noindent
\textbf{Framework Overview.}
Based on the above discussions, we propose to extract high-quality tag-style prompts from the LR image to guide the pretrained T2I model, such as stable diffusion (SD) \cite{rombach2022high}, for producing semantics-preserved Real-ISR results. The framework of our proposed method, namely \textbf{Se}mantic-awar\textbf{e} \textbf{SR} (\textbf{SeeSR}), is shown in Fig. \ref{fig:method}. The training of SeeSR goes through two stages. In the first stage (Fig. \ref{fig:method}(a)), we learn a  degradation-aware prompt extractor (DAPE), which consists of an image encoder and a tagging head. It is expected that both the feature representations and tagging outputs of the LR image can be as close as possible to that of the corresponding HR image by using the original tag model. The learned DAPE is copied to the second stage (Fig. \ref{fig:method}(b)) to extract the feature representations and tags (as text prompts) from the input LR image, which serve as control signals over the pretrained T2I model to generate visually pleasing and semantically correct Real-ISR results. During inference, only the second stage is needed to process the input image. Fig. \ref{fig:method}(c) illustrates the collaborative interplay between the image branch, feature representation branch, and text prompt branch in governing the pretrained T2I model.

\begin{table}[t]
\centering
\caption{Comparison of different prompt styles.}
\small
\begin{tabular}{cccc}
    \toprule
     & \makecell{Rich \\ Objects}  & \makecell{Concise \\ Description} & \makecell{Degradation \\ Aware} \\
    \midrule
    Classification-style & \textcolor{red}{\XSolidBrush} & \textcolor{green_c}{\Checkmark} & \textcolor{green_c}{\Checkmark} \\
    Caption-style & \textcolor{green_c}{\Checkmark} &\textcolor{red}{\XSolidBrush} & \textcolor{red}{\XSolidBrush} \\
    Tag-style     & \textcolor{green_c}{\Checkmark} \textcolor{green_c}{\Checkmark} & \textcolor{green_c}{\Checkmark} & \textcolor{red}{\XSolidBrush} \\ \midrule
    Our DAPE & \textcolor{green_c}{\Checkmark} \textcolor{green_c}{\Checkmark} & \textcolor{green_c}{\Checkmark} & \textcolor{green_c}{\Checkmark} \\ 
    \bottomrule
\end{tabular}
\label{tab:prompt_style}
\vspace{-2.0em}
\end{table}

\vspace{-1mm}


\subsection{Degradation-Aware Prompt Extractor}
\label{sec:method_1}
 The DAPE is fine-tuned from a pretrained tag model, \ie, RAM \cite{2023ram}. As depicted in Fig. \ref{fig:method}(a), the HR image $x$ goes through a frozen tag model to output representation embedding $f_{x}^{rep}$ and logits embedding $f_{x}^{logits}$ as anchor points to supervise the training of DAPE. LR images $y$ are obtained by applying random degradations to $x$, and they are fed into the trainable image encoder and tagging head. To make DAPE robust to image degradation, we force the representation embedding and logits embedding from the LR branch to be close to that of the HR branch. The training objective is as follows:
\begin{equation}
\begin{aligned}
   \mathcal{L}_{DAPE} = 
\mathcal{L}_{r}(f_{y}^{rep}, f_{x}^{rep}) + \lambda\mathcal{L}_{l}(f_{y}^{logits}, f_{x}^{logits}),
\end{aligned}
\label{eq: dape}
\end{equation}
where $\lambda$ is a balance parameter, $f_{y}^{rep}$ and $f_{y}^{logits}$ are the representation embedding and logits embedding from LR branch. $\mathcal{L}_{r}$ is the mean squared error (MSE) loss, while $\mathcal{L}_{l}$ is the cross-entropy loss \cite{he2016deep}. By aligning the outputs from LR and HR branches, DAPE is learned to predict high-quality semantic prompts from corrupted image inputs. 

Once trained, DAPE undertakes the crucial role of extracting reliable semantic prompts from the LR images. The prompts can be classified into two categories: hard prompts (\ie, tag texts from the tagging head) and soft prompts (\ie, representation embeddings from the image encoder). As shown in Figs. \ref{fig:method}(b) and \ref{fig:method}(c), hard prompts are directly passed to the frozen text encoder built into the T2I model to enhance its local understanding capability. The abundance of text prompts is controlled by a preset threshold. If the threshold is too high, the accuracy of predicted categories will improve but the recall rate can be affected, and vice versa.
Therefore, the soft label prompts are used to compensate for the limitations of hard prompts, which are free of the impact of threshold and avoid the low information entropy issue caused by one-hot categories \cite{hinton2015distilling}. 
\subsection{Training of SeeSR Model}
\label{sec:method_2}
Fig. \ref{fig:method}(c) illustrates the detailed structure of the controlled T2I diffusion model. 
Given the successful application of ControlNet \cite{zhang2023adding} in conditional image generation, we utilize it as the controller of the T2I model for Real-ISR purpose. In specific, we clone the encoder of the Unet in pre-trained SD model as a trainable copy to initialize the ControlNet. 
To incorporate soft prompts into the diffusion process, we adopt the cross-attention mechanism proposed in PASD \cite{yang2023pixel} to learn semantic guidance. The representation cross-attention (RCA) modules are added to the Unet and placed after the text cross-attention (TCA) modules. Note that the randomly initialized RCA modules are cloned simultaneously with the encoder. In addition to the text branch and representation branch, the image branch also plays a role in reconstructing the desired HR image. We pass the LR images through a trainable image encoder to obtain the LR latent, which is input to ControlNet. The structure of trainable image encoder is kept the same as that in \cite{zhang2023adding}.

The training process of the SeeSR model is as follows. The latent representation of an HR image is obtained by the encoder of pretrained VAE \cite{rombach2022high}, denoted as $z_{0}$. The diffusion process progressively introduces noise to $z_{0}$, resulting in a noisy latent $z_{t}$, where $t$ represents the randomly sampled diffusion step. With the diffusion step $t$, LR latent $z_{lr}$, hard prompts $p_{h}$ and soft prompts $p_{s}$, we train our SeeSR network, denoted as $\epsilon_{\theta}$, to estimate the noise added to the noisy latent $z_{t}$. The optimization objective is:
\begin{equation}
\begin{aligned}
\mathcal{L}=  \mathbb{E}_{z_0, z_{lr}, t, p_{h}, p_{s}, \epsilon \sim \mathcal{N}}\left[\left\|\epsilon-\epsilon_\theta\left(\mathbf{z}_t, z_{lr}, t, p_{h}, p_{s}\right)\right\|_2^2\right].
\end{aligned}
\end{equation}
For saving the training cost, we freeze the parameters of the SD model while training solely on the newly added modules, including the image encoder, the ControlNet and the RCA modules within the Unet.

\subsection{LR Embedding in Inference}
\label{sec:method_4}
\begin{figure}[htbp]
  \centering
  \includegraphics[scale=0.22]{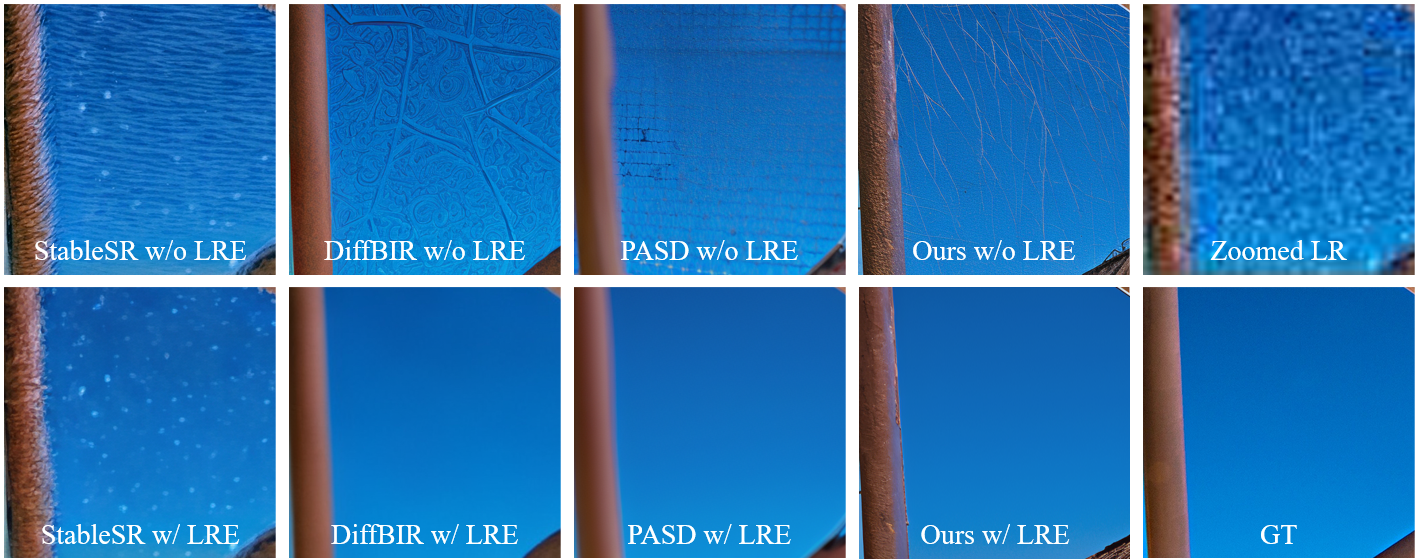}
  \caption{Effectiveness of the LR embedding (LRE) strategy in alleviating the discrepancy between training and inference of SD-based Real-ISR methods \cite{wang2023exploiting,yang2023pixel,lin2023diffbir}. Top row: results without using LRE. Bottom row: results with LRE. We see that many falsely generated details in the sky area are removed.}
  \label{fig:elr}
  \vspace{-3mm}
\end{figure}

The pretrained T2I models such as SD, during their training phase, do not completely convert the images into random Gaussian noises. However, during the inference process, most of existing SD-based Real-ISR methods \cite{wang2023exploiting,yang2023pixel,lin2023diffbir} take a random Gaussian noise as their start point, leading to a discrepancy on the noise handling procedure between training and inference \cite{lin2023common}. In the Real-ISR task, we observe that this discrepancy can confuse the model to perceive degradation as content to be enhanced, particularly in smooth regions such as the sky, as shown in the top row of Fig. \ref{fig:elr}. 
To address this issue, we propose to directly embed the LR latent into the initial random Gaussian noise according to the training noise scheduler. This strategy is applicable to most of the SD-based Real-ISR methods \cite{wang2023exploiting,yang2023pixel,lin2023diffbir}. As shown in the bottom row of Fig. \ref{fig:elr}, the proposed LR embedding (LRE) strategy alleviate much the inconsistency between training and inference, providing a more faithful start point for the diffusion model and consequently suppressing much the artifacts in the sky region. 
Note that all experiments of SeeSR in the subsequent sections utilize the LRE strategy by default.
\vspace{-3mm}


\begin{table*}[]\footnotesize
\centering
\caption{Quantitative comparison with state-of-the-art methods on both synthetic and real-world benchmarks. The best and second best results of each metric are highlighted in
\textcolor{red}{\textbf{red}} and \textcolor{blue}{\underline{blue}}, respectively. LDM is not tested on RealLR200 because the related codebase does
not provide tiled functionality, which results in the issue of out-of-memory when testing on higher-resolution inputs.}
\resizebox{\textwidth}{!}{
\begin{tabular}{@{}c|c|ccccc|cccccc@{}}
\toprule
Datasets                                                               & Metrics & BSRGAN \cite{zhang2021designing} & \begin{tabular}[c]{@{}c@{}}Real- \cite{wang2021real}\\ ESRGAN\end{tabular} & LDL \cite{liang2022details}   & DASR \cite{liang2022efficient}   & FeMaSR \cite{chen2022femasr} & LDM \cite{rombach2022high}   & StableSR \cite{wang2023exploiting} & ResShift \cite{yue2023resshift} & PASD \cite{yang2023pixel}   & DiffBIR \cite{lin2023diffbir} & SeeSR  \\ \midrule
\multirow{9}{*}{\begin{tabular}[c]{@{}c@{}}\textit{DIV2K-Val}\end{tabular}} & PSNR $\uparrow$   & \textcolor{blue}{\underline{21.87}}  & \textcolor{red}{\textbf{21.94}}                                                & 21.52  & 21.72  & 20.85  & 21.26  & 20.84    & 21.75    & 20.77  & 20.94   & 21.19  \\
                                                                            & SSIM $\uparrow$   & 0.5539 & \textcolor{red}{\textbf{0.5736}}                                                & \textcolor{blue}{\underline{0.5690}} & 0.5536 & 0.5163 & 0.5239 & 0.4887   & 0.5422   & 0.4958 & 0.4938  & 0.5386 \\
                                                                            & LPIPS $\downarrow$  & 0.4136 & \textcolor{blue}{\underline{0.3868}}                                                 & 0.3995 & 0.4266 & 0.3973 & 0.4154 & 0.4055   & 0.4284   & 0.4410 & 0.4270  & \textcolor{red}{\textbf{0.3843}} \\
                                                                            & DISTS $\downarrow$  & 0.2737 & 0.2601                                                 & 0.2688 & 0.2688 & \textcolor{blue}{\underline{0.2428}} & 0.2500 & 0.2542   & 0.2606   & 0.2538 & 0.2471  & \textcolor{red}{\textbf{0.2257}} \\
                                                                            & FID $\downarrow$    & 64.28  & 53.46                                                  & 58.94  & 67.22  & 53.70  & 41.93  & \textcolor{blue}{\underline{36.57}}    & 55.77    & 40.77  & 40.42   & \textcolor{red}{\textbf{31.93}}  \\
                                                                            & NIQE $\downarrow$   & 4.7615 & 4.9209                                                 & 5.0249 & 4.8596 & \textcolor{red}{\textbf{4.5726}} & 6.4667 & \textcolor{blue}{\underline{4.6551}}   & 6.9731   & 4.8328 & 4.7211  & 4.9275 \\
                                                                            & MANIQA $\uparrow$ & 0.4834 & 0.5251                                                 & 0.5127 & 0.4346 & 0.4869 & 0.5237 & 0.5914   & 0.5232   & 0.6049 & \textcolor{red}{\textbf{0.6205}}  & \textcolor{blue}{\underline{0.6198}} \\
                                                                            & MUSIQ $\uparrow$  & 59.11  & 58.64                                                  & 57.90  & 54.22  & 58.10  & 56.52  & 62.95    & 58.23    & \textcolor{blue}{\underline{66.85}}  & 65.23   & \textcolor{red}{\textbf{68.33}}  \\
                                                                            & CLIPIQA $\uparrow$& 0.5183 & 0.5424                                                 & 0.5313 & 0.5241 & 0.5597 & 0.5695 & 0.6486   & 0.5948   & \textcolor{blue}{\underline{0.6799}} & 0.6664  & \textcolor{red}{\textbf{0.6946}} \\ \midrule
\multirow{9}{*}{\begin{tabular}[c]{@{}c@{}}\textit{RealSR}\end{tabular}}     & PSNR $\uparrow$   & \textcolor{blue}{\underline{26.39}}  & 25.69                                                  & 25.28  & \textcolor{red}{\textbf{27.02}}  & 25.07  & 25.48  & 24.70    & 26.31    & 24.29  & 24.77   & 25.18  \\
                                                                            & SSIM $\uparrow$   & \textcolor{blue}{\underline{0.7654}} & 0.7616                                                 & 0.7567 & \textcolor{red}{\textbf{0.7708}} & 0.7358 & 0.7148 & 0.7085   & 0.7421   & 0.6630 & 0.6572  & 0.7216 \\
                                                                            & LPIPS $\downarrow$  & \textcolor{red}{\textbf{0.2670}} & \textcolor{blue}{\underline{0.2727}}                                                 & 0.2766 & 0.3151 & 0.2942 & 0.3180 & 0.3018   & 0.3460   & 0.3435 & 0.3658  & 0.3009 \\
                                                                            & DISTS $\downarrow$  & \textcolor{blue}{\underline{0.2121}} & \textcolor{red}{\textbf{0.2063}}                                                 & \textcolor{blue}{\underline{0.2121}} & 0.2207 & 0.2288 & 0.2213 & 0.2135   & 0.2498   & 0.2259 & 0.2310  & 0.2223 \\
                                                                            & FID $\downarrow$    & 141.28 & 135.18                                                 & 142.71 & 132.63 & 141.05 & 132.72 & \textcolor{blue}{\underline{128.51}}   & 141.71   & 129.76 & 128.99  & \textcolor{red}{\textbf{125.55}} \\
                                                                            & NIQE $\downarrow$   & 5.6567 & 5.8295                                                 & 6.0024 & 6.5311 & 5.7885 & 6.5200 & 5.9122   & 7.2635   & \textcolor{red}{\textbf{5.3628}} & 5.5696  & \textcolor{blue}{\underline{5.4081}} \\
                                                                            & MANIQA $\uparrow$ & 0.5399 & 0.5487                                                 & 0.5485 & 0.3878 & 0.4865 & 0.5423 & 0.6221   & 0.5285   & \textcolor{red}{\textbf{0.6493}} & 0.6253  & \textcolor{blue}{\underline{0.6442}} \\
                                                                            & MUSIQ $\uparrow$  & 63.21  & 60.18                                                  & 60.82  & 40.79  & 58.95  & 58.81  & 65.78    & 58.43    & \textcolor{blue}{\underline{68.69}}  & 64.85   & \textcolor{red}{\textbf{69.77}}  \\
                                                                            & CLIPIQA $\uparrow$& 0.5001 & 0.4449                                                 & 0.4477 & 0.3121 & 0.5270 & 0.5709 & 0.6178   & 0.5444   & \textcolor{blue}{\underline{0.6590}} & 0.6386  & \textcolor{red}{\textbf{0.6612}} \\ \midrule
\multirow{9}{*}{\begin{tabular}[c]{@{}c@{}}\textit{DrealSR}\end{tabular}}     & PSNR $\uparrow$   & \textcolor{blue}{\underline{28.75}}  & 28.64                                                  & 28.21  & \textcolor{red}{\textbf{29.77}}  & 26.90  & 27.98  & 28.13    & 28.46    & 27.00  & 26.76   & 28.17  \\
                                                                            & SSIM $\uparrow$   & 0.8031 & 0.8053                                                 & \textcolor{blue}{\underline{0.8126}} & \textcolor{red}{\textbf{0.8264}} & 0.7572 & 0.7453 & 0.7542   & 0.7673   & 0.7084 & 0.6576  & 0.7691 \\
                                                                            & LPIPS $\downarrow$  & 0.2883 & \textcolor{blue}{\underline{0.2847}}                                                 & \textcolor{red}{\textbf{0.2815}} & 0.3126 & 0.3169 & 0.3405 & 0.3315   & 0.4006   & 0.3931 & 0.4599  & 0.3189 \\
                                                                            & DISTS $\downarrow$  & 0.2142 & \textcolor{red}{\textbf{0.2089}}                                                 & \textcolor{blue}{\underline{0.2132}} & 0.2271 & 0.2235 & 0.2259 & 0.2263   & 0.2656   & 0.2515 & 0.2749  & 0.2315 \\
                                                                            & FID  $\downarrow$   & 155.63 & \textcolor{blue}{\underline{147.62}}                                                 & 155.53 & 155.58 & 157.78 & 156.01 & 148.98   & 172.26   & 159.24 & 166.79  & \textcolor{red}{\textbf{147.39}} \\
                                                                            & NIQE $\downarrow$   & 6.5192 & 6.6928                                                 & 7.1298 & 7.6039 & \textcolor{blue}{\underline{5.9073}} & 7.1677 & 6.5354   & 8.1249   & \textcolor{red}{\textbf{5.8595}} & 6.2935  & 6.3967 \\
                                                                            & MANIQA $\uparrow$ & 0.4878 & 0.4907                                                 & 0.4914 & 0.3879 & 0.4420 & 0.5043 & 0.5591   & 0.4586   & 0.5850 & \textcolor{blue}{\underline{0.5923}}  & \textcolor{red}{\textbf{0.6042}} \\
                                                                            & MUSIQ $\uparrow$  & 57.14  & 54.18                                                  & 53.85  & 42.23  & 53.74  & 53.73  & 58.42    & 50.60    & \textcolor{blue}{\underline{64.81}}  & 61.19   & \textcolor{red}{\textbf{64.93}}  \\
                                                                            & CLIPIQA $\uparrow$& 0.4915 & 0.4422                                                 & 0.4310 & 0.3684 & 0.5464 & 0.5706 & 0.6206   & 0.5342   & \textcolor{blue}{\underline{0.6773}} & 0.6346  & \textcolor{red}{\textbf{0.6804}} \\ \midrule                                                                          
\multirow{4}{*}{\begin{tabular}[c]{@{}c@{}}\textit{RealLR200}\end{tabular}}  & NIQE $\downarrow$   & 4.3817 & 4.2048                                                 & 4.3845 & 4.3360 & 4.6357 & -      & 4.2516   & 6.2878   & \textcolor{blue}{\underline{4.1715}} & 4.9330  & \textcolor{red}{\textbf{4.1620}} \\
                                                                            & MANIQA $\uparrow$ & 0.5462 & 0.5582                                                 & 0.5519 & 0.4877 & 0.5295 & -      & 0.5841   & 0.5417   & \textcolor{blue}{\underline{0.6066}} & 0.5902  & \textcolor{red}{\textbf{0.6254}} \\
                                                                            & MUSIQ $\uparrow$  & 64.87  & 62.94                                                  & 63.11  & 55.67  & 64.14  & -      & 63.30    & 60.18    & \textcolor{blue}{\underline{68.20}}  & 62.06   & \textcolor{red}{\textbf{69.71}}  \\
                                                                            & CLIPIQA $\uparrow$& 0.5679 & 0.5389                                                 & 0.5326 & 0.4659 & 0.6522 & -      & 0.6068   & 0.6486   & \textcolor{blue}{\underline{0.6797}} & 0.6509  & \textcolor{red}{\textbf{0.6813}} \\ \bottomrule
\end{tabular}

}
\label{tab:methods}
\vspace{-2mm}
\end{table*}

\section{Experiments}
\label{sec:exp}

Following previous works \cite{zhang2021designing, wang2021real}, we focus on the challenging $\times$4 Real-ISR tasks, while the proposed method can be applied to other scaling factors. Furthermore, we evaluate the semantic restoration capability of SeeSR and other Real-ISR methods on the well-known COCO dataset \cite{coco}. 
\subsection{Experimental Settings}
\noindent
\textbf{Training Datasets.} We train SeeSR on LSDIR \cite{li2023lsdir} and the first 10K face images from FFHQ \cite{ffhq}. The degradation pipeline of Real-ESRGAN \cite{wang2021real} is used to synthesize LR-HR training pairs. 

\noindent
\textbf{Test Datasets.} We employ the following test datasets to comprehensively evaluate SeeSR. (1) First, we randomly crop 3K patches (resolution: 512$\times$512) from the DIV2K validation set \cite{div2k} and degrade them using the same pipeline as that in training. We name this dataset as \textit{DIV2K-Val}. (2) We employ the two real-world datasets, \textit{RealSR} \cite{realsr} and \textit{DRealSR} \cite{drealsr}, by using the same configuration as \cite{wang2023exploiting} to center-crop the LR image  to $128\times128$ \footnote{\noindent https://huggingface.co/datasets/Iceclear/StableSR-TestSets}. (3) We build another real-world dataset, named \textit{RealLR200}, which comprises 38 LR images used in recent literature \cite{wang2021real, liang2022details, zhang2017beyond}, 47 LR images from DiffBIR \cite{lin2023diffbir}, 50 LR images from VideoLQ (the last frame of each video sequence) \cite{chan2022investigating}, and 65 LR images collected from the internet by ourselves.

\noindent
\textbf{Implementation Details.} There exist many efficient methods  \cite{lora, dettmers2024qlora, zhang2024dual} of fine-tuning. We utilize the well-known LORA (r = 8) method \cite{lora}  to fine-tune the entire DAPE module from RAM \cite{2023ram} for 20K iterations. The batch size and the learning rate are set to 32 and $10^{-4}$, respectively. The SD 2-base\footnote{\noindent https://huggingface.co/stabilityai/stable-diffusion-2-base} is used as the pretrained T2I model. The whole controlled T2I model is trained for 150K iterations with the Adam \cite{adam} optimizer, where the batch size and learning rate are respectively set to 192 and $5\times10^{-5}$. The training process is conducted on 512$\times$512 resolution images with 8 NVIDIA Tesla 32G-V100 GPUs. During inference, we adopt the spaced DDPM sampling \cite{nichol2021improved} with 50 timesteps. $\lambda$ in Eq. (\ref{eq: dape}) is set to 1.

\noindent
\textbf{Evaluation Metrics.} In order to provide a comprehensive and holistic assessment of the performance of different methods, we employ a range of reference and no-reference metrics.
PSNR and SSIM \cite{ssim} (calculated on the Y channel in YCbCr space) are reference-based fidelity measures, while LPIPS\footnote{\noindent We use LPIPS-Alex by default.} \cite{lpips}, DISTS \cite{dists} are reference-based perceptual quality measures. FID \cite{fid} evaluates the distance of distributions between original and restored images. NIQE \cite{niqe}, MANIQA \cite{maniqa}, MUSIQ \cite{musiq}, and CLIPIQA \cite{clipiqa} are no-reference image quality measures. 

\begin{figure*}[t]
  \centering
  \includegraphics[scale=0.155]{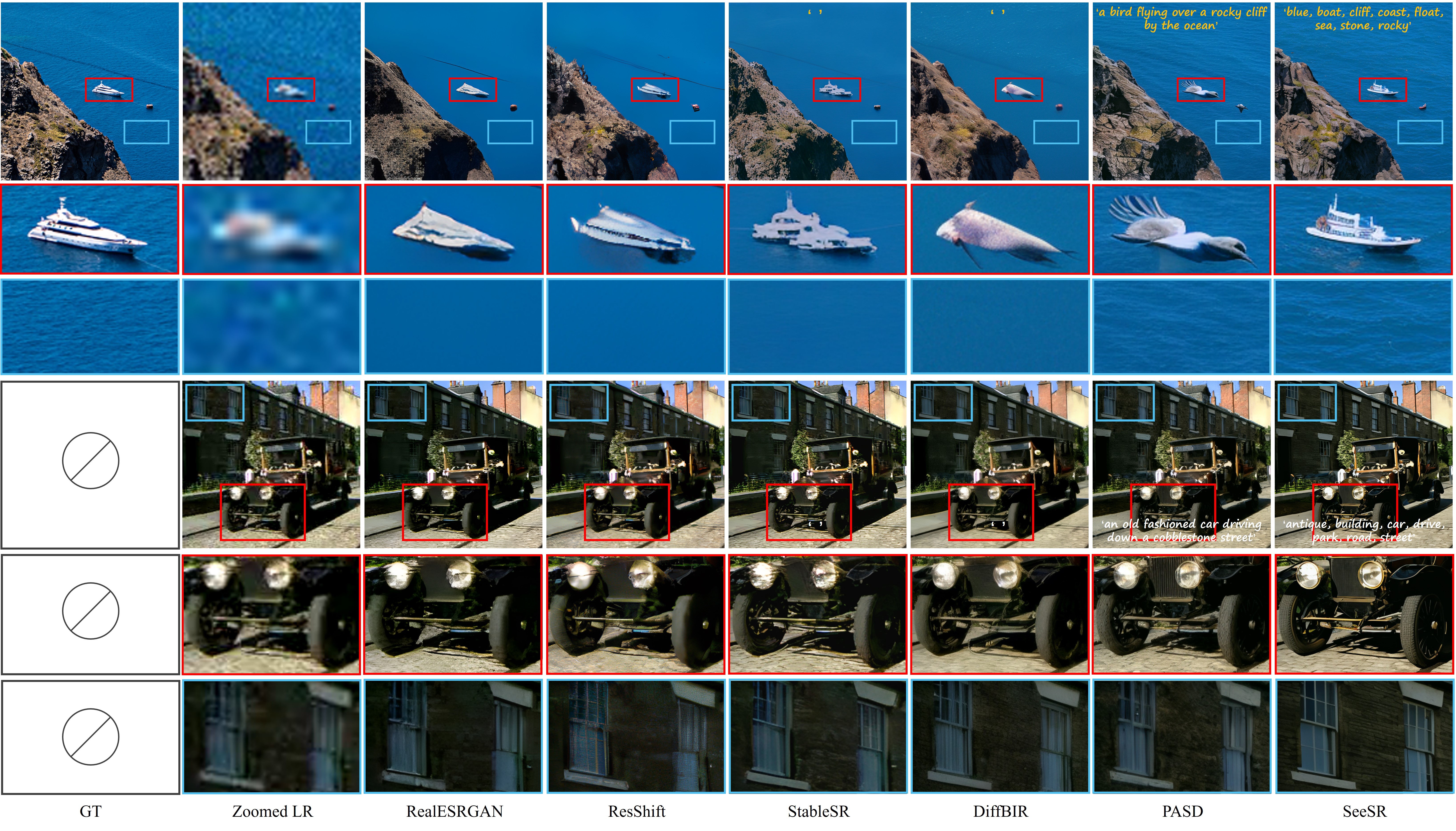}
\vspace{-3mm}
  \caption{Qualitative comparisons of different Real-ISR methods. Please zoom in for a better view.}
  \label{fig:data_sys_real}
  \vspace{-3mm}
\end{figure*}

\noindent
\textbf{Compared Methods.} 
We compare our SeeSR with several state-of-the-art Real-ISR methods, which can be categorized into two groups. The first group consists of GAN-based methods, including BSRGAN \cite{zhang2021designing}, Real-ESRGAN \cite{wang2021real}, LDL \cite{liang2022details}, FeMaSR \cite{chen2022femasr} and DASR \cite{liang2022efficient}. The second group consists of recent diffusion-based methods, including LDM \cite{rombach2022high},  StableSR \cite{wang2023exploiting}, ResShift \cite{yue2023resshift}, PASD \cite{yang2023pixel}, and DiffBIR \cite{lin2023diffbir}. We use the publicly released codes and models of the competing methods for testing. 

\subsection{Comparison with State-of-the-Arts}
\label{sec:exp_comp}

\noindent
\textbf{Quantitative Comparisons.}
We first show the quantitative comparison on the four synthetic and real-world datasets in Table \ref{tab:methods}. We can have the following observations. (1) First, our SeeSR consistently achieves the best scores in FID, CLIPIQA and MUSIQ across all the four datasets. (2) Second, SeeSR achieves the best LPIPS and DISTS scores on \textit{DIV2K-Val}, surpassing the corresponding second-best methods by more than 0.6\% and 7.0\%, respectively. (3) GAN-based methods achieve better PSNR/SSIM scores than DM-based methods. This is mainly because DM-based methods can generate more realistic details, which however sacrifice the fidelity. (4) BSRGAN, Real-ESRGAN and LDL show advantages in terms of reference perceptual metrics LPIPS/DISTS, but they perform poorer in no-reference perceptual metrics such as CLIPIQA, MUSIQ and MANIQA. This is also because DM-based methods will generate some structures and textures that may not match the GTs, making them disadvantageous in full-reference metrics. Overall, compared with other DM-based methods, our SeeSR achieves better no-reference metric scores, while keep competitive full-reference measures.

\begin{table*}[ht]\footnotesize
\centering
\caption{The comparison of semantic restoration performance among different Real-ISR methods.}
\vspace{-2mm}
\resizebox{\textwidth}{!}{

\begin{tabular}{@{}c|cc|ccccc|cccccc@{}}
\toprule
Metrics  & GT & Zoomed LR & BSRGAN & \begin{tabular}[c]{@{}c@{}}Real-\\ ESRGAN\end{tabular} & LDL & DASR & FeMaSR & LDM & StableSR & ResShift & PASD & DiffBIR & SeeSR \\ \midrule
Panoptic Segmentation (PQ) & 52.5   &  9  &  16.2      &  19.4     &   17.8  &  15.5    &     15.6   &   18.7  &      26.8    &   21.4    &   23.7   &    27.2     &   \textbf{30.0}    \\
Object Detection (AP) &  49.1  &  5  &   10.5     &   13.1 &   11.9  &  9.9    &   10.1     &   11.4  &     18.3     &   14.3     &  15.5    &    18.9     &   \textbf{21.1}    \\
Instance Segmentation (AP) &  43.8  & 4   &   9.2     &    11.4  &   10.3  &   8.6   &    8.8    &  9.9   &   16.2       &    12.4    &   13.4   &    16.5     &    \textbf{18.5}   \\
Semantic Segmentation (mIOU) &  62.0  &  12  &  21.7      &  26.0  &   24.2  &  20.5    &   20.4     &  25.3   &    34.5      &    30.4      &   33.3   &   37.7      &   \textbf{41.3}    \\ \bottomrule
\end{tabular}
}
\label{tab: semantic}
\vspace{-1mm}
\end{table*}

\noindent
\textbf{Qualitative Comparisons.}
Fig. \ref{fig:data_sys_real} presents visual comparisons on synthetic and real-world images, respectively. We also provide the caption-style prompts predicted by PASD and the tag-style prompts predicted by SeeSR. As illustrated in the first case of Fig. \ref{fig:data_sys_real}, Real-ESRGAN fails to reconstruct the details of the ship, which suffers from severe degradation. ResShift is unable to reconstruct the details of the ship either due to the lack of pretrained image priors. Without using textual cues, the output of StableSR is semantically ambiguous between ship and building. As a result of the ambiguous output of its degradation removal stage, DiffBIR wrongly reconstructs the ship into fish. Additionally, the aforementioned methods smooth out the ripples of the sea, reducing the liveliness of the reconstructed images. The caption model of PASD outputs text prompts with semantic errors, wrongly generating a bird. In comparison, our well-trained DAPE module in SeeSR can still provide accurate prompt even with strong degradation, aiding SeeSR to generate semantically-accurate and details-rich results.  

Similar conclusions can be drawn from the second case with a real-world LR image. Real-ESRGAN and ResShift generate limited and unnatural details. Concurrently, although StableSR and DiffBIR utilize pretrained priors, the detailed structure of the muntins and bricks are not orderly due to the absence of text prompt guidance. The text prompts of PASD miss the building, which lead to limited semantic details in the corresponding scene. In contrast, SeeSR predicts most of the tags, including the building, so that the semantic details of the windows and walls are well recovered. Besides, thanks to the soft prompting mechanism, even the predicted tags do not explicitly include the ``tires'', SeeSR still produces tires with rich textures.

\noindent 
\textbf{User Study.}
To further validate the effectiveness of our method, we conduct separate user studies on synthetic data and real data. On synthetic data, inspired by SR3 \cite{sr3}, participants were presented with an LR image placed between two HR images each time: one is the GT and another is the Real-ISR output by one model. They were asked to determine \textit{`Which HR image better corresponds to the LR image?'} When making decision, participants were asked to consider two factors: the perceptual quality of the HR image and its semantic similarity to the LR image. Then the confusion rates can be calculated, which indicate the participants' preference to the GT or the Real-ISR output. 
On real-world data, participants were presented with an LR image alongside all Real-ISR outputs, and they were asked to answer \textit{`Which image is the best SR result of the LR image?'} In this experiment, best rates were calculated, which represent the probability of the model being selected. 

We invited 20 participants to test six representative methods (Real-ESRGAN, StableSR, PASD, DiffBIR, ResShift and SeeSR). There are 16 synthetic test sets and 16 real-world test sets. The synthetic data are randomly sampled from \textit{DIV2K-Val}, and the real-world data are randomly sampled from \textit{RealLR200}. Each of the 20 participants was asked to make 112 selections ($16\times6+16$). As shown in Table \ref{tab: user_study}, our SeeSR significantly outperforms others in terms of selection rate on both synthetic and real data. In the user study on synthetic data, the SR results of all models cannot compete with the GT, while our SeeSR achieves a confusion rate of 38.6\%, which is three times higher than the second-ranked method. This implies that there is still enough room to improve for the Real-ISR methods. In the user study on real-world data where there is no GT, our method achieves a best selection rate of 57.1\%, approximately 3.5 times higher than the second-ranked method. 

\begin{table}[t] \scriptsize 
\centering
\caption{Results of user study on synthetic and real-world data.}

\begin{tabular}{c|c|c}
\hline
Methods     & \begin{tabular}[c]{@{}c@{}}Confusion rates \\ on synthetic data\end{tabular} & \begin{tabular}[c]{@{}c@{}}Best rates\\  on real-world data\end{tabular} \\ \hline
Real-ESRGAN &     5.4\%                                            &      0\%                                                                    \\
StableSR    &     5.4\%                                                                         &     14.3\%                                                                      \\
ResShift        &     3.6\%                                                                         &   0\%                                                                       \\
PASD    &             10.7\%                                                                 &      13.4\%                                                                    \\
DiffBIR     &         12.5\%                                                                     &   15.2\%                                                                       \\
SeeSR       &         \textbf{38.6\%}                                               &   \textbf{57.1\%}                                                                       \\ \hline
\end{tabular}
\label{tab: user_study}
\vspace{-3mm}
\end{table}

\vspace{-1mm}
\subsection{Semantics Preservation Test}
To further validate our model's ability to preserve semantic fidelity, we conduct detection \cite{redmon2016you, li2023one} and segmentation tasks \cite{he2017mask, li2023mdqe, li2024univs} on the Real-ISR output images. We resize the original images from COCO-Val (5K images) ~\cite{coco} to $512\times512$ as GT, and then degrade them to generate LR images as in training. We employ OpenSeeD~\cite{openseed} trained on COCO  as the detector and segmentor since it is a strong transformer-based unified model for segmentation and detection tasks. As shown in Table \ref{tab: semantic}, compared to Zoomed LR, SeeSR achieves a remarkable $3\sim4$ times improvement in all four tasks, surpassing all existing Real-ISR methods and showcasing its strong semantics preservation capability.

\vspace{-1mm}
\section{Conclusion and Limitation}
\label{sec:conclusion and Limitation}

We proposed SeeSR, a Real-ISR method that utilizes semantic prompts to enhance the generative capability of pretrained T2I diffusion models. Through exploring the impact of different styles of text prompts on the generated results, we found that the image tags can greatly enhance the local perception ability of the T2I model. However, the tags are susceptible to complex image degradation, and they are influenced by manually set thresholds. Therefore, we proposed DAPE, which minimizes the influence of image degradation on semantic prompts and simultaneously outputs soft and hard semantic prompts to guide the diffusion process in image super-resolution. Furthermore, to address the adverse effects of training-test inconsistency in diffusion models, we proposed a simple yet effective LRE strategy, which embeds LR latent at the starting point of diffusion process, avoiding the generation of artifacts in smooth areas.  
Our work made a step towards better leveraging generative priors to synthesize semantically correct Real-ISR images, as demonstrated in our extensive experiments.

There are some limitations of SeeSR. First of all, DAPE may predict incorrect tags for heavily degraded images, resulting in wrongly restored objects. Second, the alignment between tags and regions in the LR image can be inaccurate in cases of severe degradation. Providing extra mask information can help alleviate this issue. Third, as in other SD-based methods, SeeSR encounters challenges in reconstructing small-scale scene text images.

{
    \small
    \bibliographystyle{ieeenat_fullname}
    \bibliography{main}
}



\end{document}